\documentclass[12pt]{spieman}  
\usepackage{amsmath,amsfonts,amssymb}
\usepackage{graphicx}
\usepackage{setspace}
\usepackage{tocloft}
\usepackage{subfig}
\usepackage{xcolor}
\usepackage{pifont}
\title{Self-Supervised Learning for Interventional Image Analytics: Towards Robust Device Trackers}

\author[a,b,*]{Saahil Islam}
\author[c]{Venkatesh N. Murthy}
\author[b]{Dominik Neumann}
\author[a,b]{Badhan Kumar Das}
\author[c]{Puneet Sharma}
\author[a]{Andreas Maier}
\author[c]{Dorin Comaniciu}
\author[b]{Florin C. Ghesu}
\affil[a]{Pattern Recognition Lab, Friedrich-Alexander-Universität Erlangen-Nürnberg, 91058 Erlangen, Germany}
\affil[b]{Digital Technology and Innovation, Siemens Healthineers, Erlangen, Germany}
\affil[c]{Digital Technology and Innovation, Siemens Healthineers, Princeton, NJ, USA}

\cftpagenumbersoff{figure}
\cftpagenumbersoff{table} 
\begin{document} 
\maketitle

\begin{abstract}\\
\textbf{Purpose:} An accurate detection and tracking of devices such as guiding catheters in live X-ray image acquisitions is an essential prerequisite for endovascular cardiac interventions. This information is leveraged for procedural guidance, e.g., directing stent placements. To ensure procedural safety and efficacy, there is a need for high robustness / no failures during tracking. To achieve that, one needs to efficiently tackle challenges, such as: device obscuration by contrast agent or other external devices or wires, changes in field-of-view or acquisition angle, as well as the continuous movement due to cardiac and respiratory motion. \\
\textbf{Approach:} To overcome the aforementioned challenges, we propose a novel approach to learn spatio-temporal features from a very large data cohort of over 16 million interventional X-ray frames using self-supervision for image sequence data. Our approach is based on a masked image modeling technique that leverages frame interpolation based reconstruction to learn fine inter-frame temporal correspondences. The features encoded in the resulting model are fine-tuned downstream in a light-weight model. \\
\textbf{Results:} Our approach achieves state-of-the-art performance and in particular robustness compared to ultra optimized reference solutions (that use multi-stage feature fusion, multi-task and flow regularization). The experiments show that our method achieves 66.31\% reduction in maximum tracking error against reference solutions (23.20\% when flow regularization is used); achieving a success score of 97.95\% at a 3$\times$ faster inference speed of 42 frames-per-second (on GPU). In addition, we achieve a 20\% reduction in the standard deviation of errors, which indicates much more stable tracking performance. \\
\textbf{Conclusions:} The proposed data-driven approach achieves superior performance particularly in robustness and speed compared to the frequently used multi-modular approaches for device tracking. The results encourage the use of our approach in various other tasks within interventional image analytics that require effective understanding of spatio-temporal semantics.

\end{abstract}


{\noindent \footnotesize\textbf{*}Saahil Islam,  \linkable{saahil.islam@fau.de} \footnote{Address all correspondence to Saahil Islam, \linkable{saahil.islam@fau.de}}}

\begin{spacing}{1}   

\section{Introduction}
The tracking of interventional devices is an important prerequisite for interventional specialists during catheterized cardiac interventions such as percutaneous coronary interventions (PCI), cardiac electrophysiology (EP), or transarterial chemoembolization (TACE)~\cite{ma2020dynamic,odening2021esc,facciorusso2015transarterial}.

Tracking the tip of the catheter as a visual guidance facilitates navigation to the desired anatomy. Furthermore, the tip of the catheter serves as an anchor point separating the catheter from the vessel structures. The anchor point can provide a basis for mapping angiography (high-dose X-ray with injected contrast agent) to fluoroscopy (low-dose X-ray), thereby reducing usage of contrast for visualizing vessels \cite{ma2020dynamic,piayda2018dynamic}. In order to co-register intravascular ultrasonography (IVUS) with angiography and perform a complete examination of the vessel, lumen, and wall structure, catheter tip tracking also offers a significant cue \cite{wang2011image,araki2015comparative,wang2013image}. 

\begin{figure}[t]
	\centering
	\includegraphics[width=0.8\columnwidth]{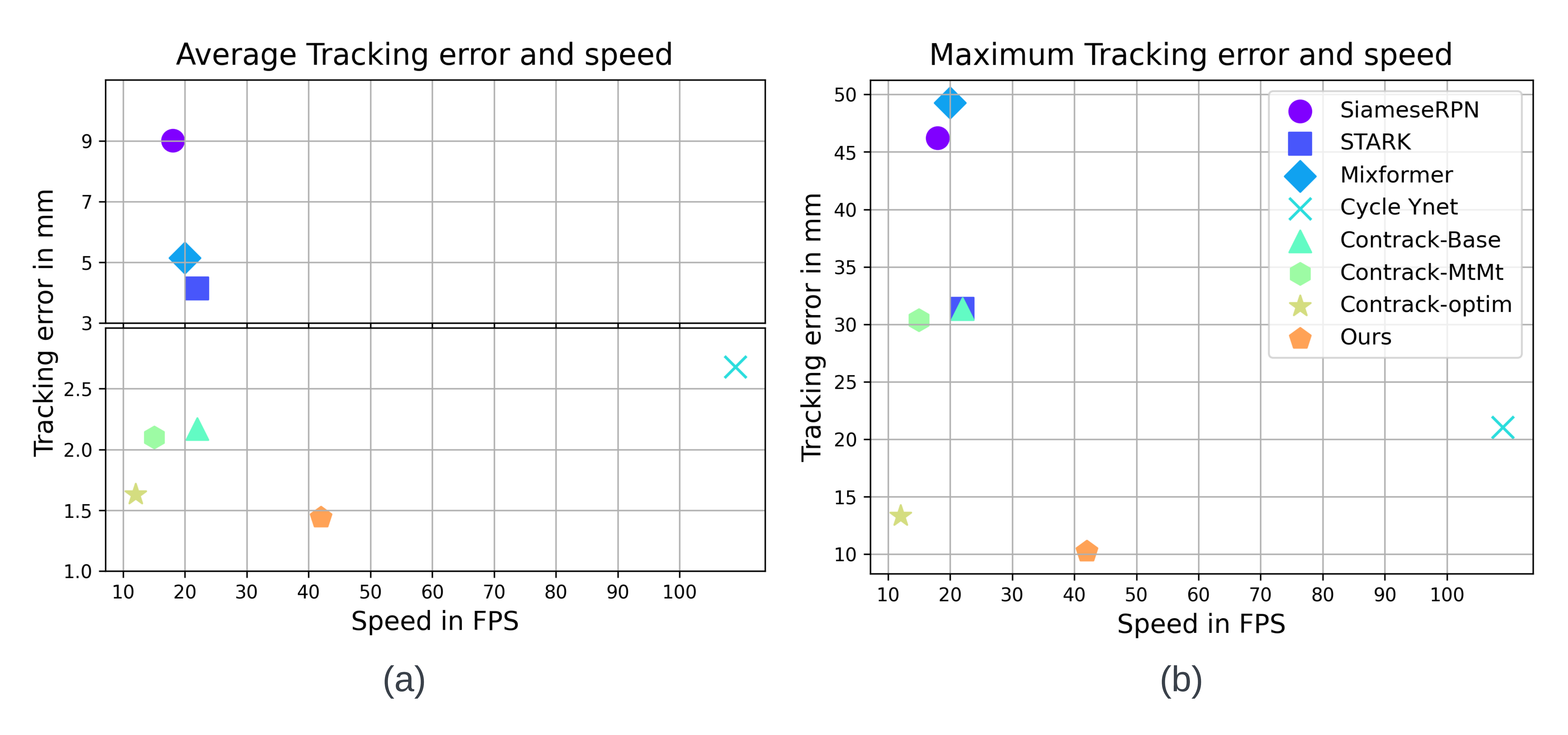}
	
    \caption[Tracking error ($\downarrow$) versus average speed ($\uparrow$): (a) showing average tracking error; and (b) showing maximum tracking error.]{Tracking error ($\downarrow$) versus average speed ($\uparrow$) for catheter tip tracking in coronary X-ray sequences acquired during procedures such as invasive coronary angiography (ICA) or percutaneous coronary intervention (PCI): (a) showing average tracking error; and (b) showing maximum tracking error. Note that the average tracking error has $2$ different scales indicated with a horizontal break-point for better visualization. Runtime is measured on a Tesla V100 GPU.}
	\label{fig:err_speed}
\end{figure}

However, tracking the tip of the catheter in X-ray images can be challenging in presence of various occlusions due to contrast agent and other devices. This is in addition to the cardiac and breathing motion of the patient. Recently, self-supervised learning methods have been developed with the aim to learn general features from unlabeled data to boost the performance in various natural sequence imaging tasks. 
Most self-supervised pretraining methods learn such features by identifying and removing inherent redundancies from sequence image data. VideoMAE \cite{tong2022videomae} conducts temporal downsampling on the pixel level followed by symmetrical masking over all the sampled frames with a high masking ratio of $90\%$. This deliberate design choice prevents the network from learning fine inter-frame correspondences. SiamMAE \cite{gupta2023siamese} improves upon this baseline by using highly asymmetric masking. However, the proposed asymmetric masking requires feeding in the first frame entirely with $0 \%$ masking which increases the compute complexity quadratically and prevents the network from learning spatio-temporal features over a longer period of time.

The space-time semantics in interventional cardiac image sequences differ from natural videos in terms of both redundancies and motion. For example, visibility may largely vary based on X-ray dosage along with varying motion based on the acquisition frame-rate, patient's breathing and cardiac motion. In angiography sequences, vessels have high structural similarity with devices such as catheters and guidewires and can gradually appear or disappear over time.



\begin{figure}[tp]
	\centering
	\includegraphics[width=0.85\columnwidth]{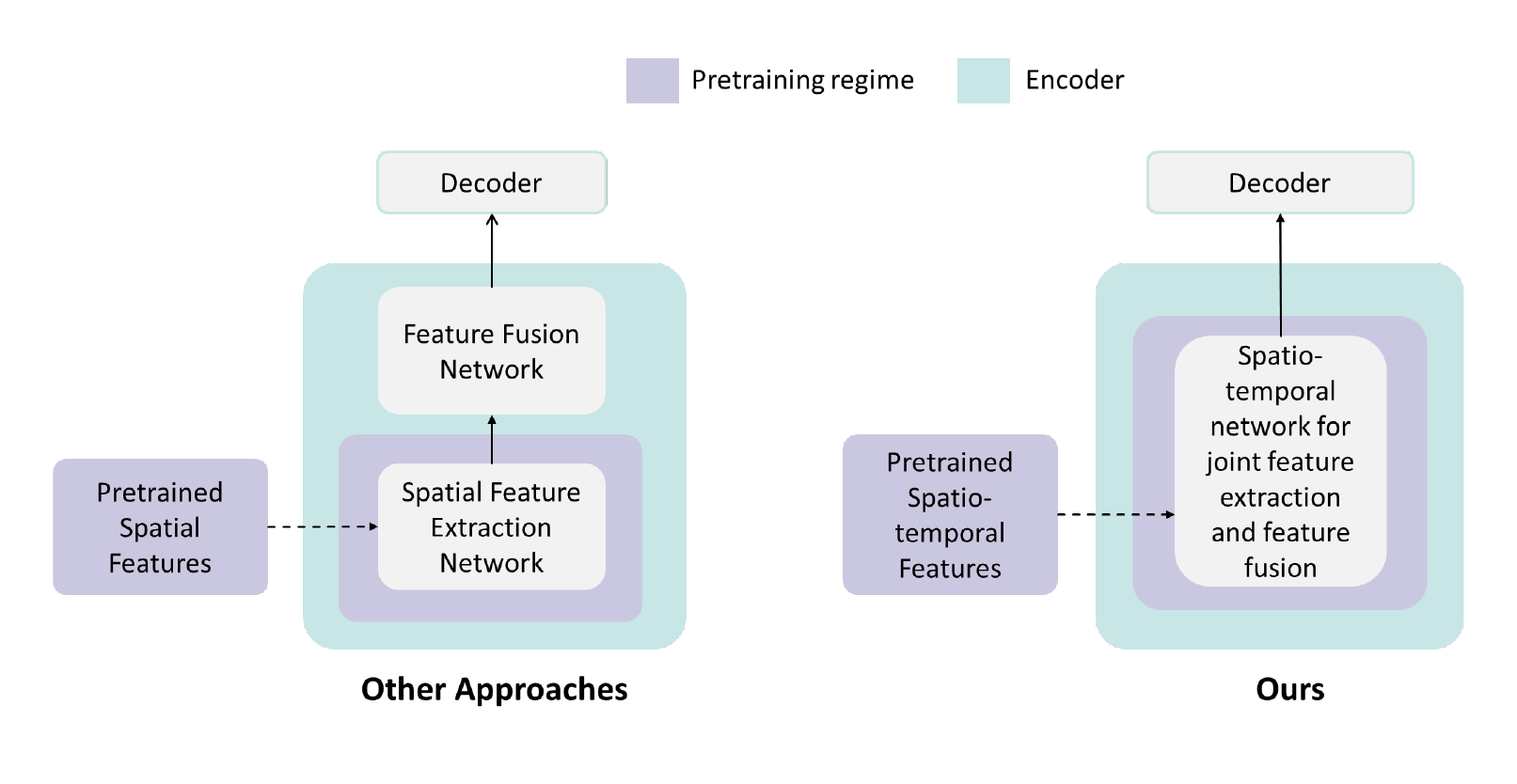}
	\caption{Overview of key differences between our approach and previous approaches for device tracking.}
	\label{fig:track_compare}
\end{figure}

To address these challenges, in this work we bring the following contributions in terms of both self-supervised pretraining and the downstream device tracking:
\begin{enumerate}
    \item We pretrain a spatio-temporal encoder on a large database of interventional cardiac X-ray sequences from over 20,000 patients (over 16,000,000 frames) for robust device tracking. 
    \item We propose a novel frame interpolation masked auto-encoder (FIMAE) to learn generalized spatio-temporal features from this dataset. The pretrained spatio-temporal features play an essential role in feature extraction and feature matching for tracking. Our pretrained features efficiently capture the underlying temporal motion needed for tracking, which is typically accomplished through highly optimized supplementary modules in other device tracking models \cite{lin2020cycle, demoustier2023contrack}.
    \item To the best of our knowledge, this is the first approach which leverages spatio-temporal pretrained features to replace a commonly used Siamese-like architecture for single object tracking.
    \item A lightweight Vision Transformer (ViT) \cite{dosovitskiy2020image} based model is designed that leverages the learned features to replace a traditional two-stage tracking encoder for feature extraction and feature fusion into one spatio-temporal encoder for highly accurate and robust real-time device tracking with an inference speed of 42 fps on a single Tesla V100 GPU (refer to Figs. \ref{fig:err_speed} and \ref{fig:track_compare}).
    \item We conduct comprehensive numerical experiments and demonstrate that our method outperforms other state-of-the-art tracking methods in robustness, accuracy and speed. 
    \item We conduct a comprehensive analysis of our model's robustness in handling long temporal sequences and demonstrate its ability to maintain consistent performance across diverse scenarios, including angiography, fluoroscopy, and sequences featuring additional obstructions caused by other devices.
   
\end{enumerate}

\section{Related Work}
\paragraph{Self-Supervised Learning} methods have been used in a variety of contexts to learn features from unlabeled data that boost the performance in downstream tasks, such as using pretext tasks \cite{doersch2015unsupervised, pathak2017learning, gidaris2018unsupervised} and contrastive learning \cite{wu2018unsupervised, he2020momentum, caron2021emerging, chen2020simple, grill2020bootstrap, chen2021exploring}. In the space of sequential image data processing (e.g., video), temporal information has been leveraged in various ways \cite{sermanet2018time, sun2019learning, han2019video, feichtenhofer2021large, recasens2021broaden, qian2021spatiotemporal, park2023self}. However, self-supervised methods based on masked image modeling (MIM) have shown significant promise recently, where the input is masked to a high percentage and fed through an encoder-decoder network to predict the missing information \cite{devlin2018bert, he2022masked, bao2021beit, xie2022simmim}. Some methods use symmetrical masking on temporally downsampled video frames to reduce space-time redundancies over a long time period \cite{feichtenhofer2022masked} \cite{tong2022videomae}. In contrast, others \cite{gupta2023siamese} use asymmetrical masking to learn inter-frame correspondence between frame pairs. However, we propose a method for both reducing space-time redundancies over a long time period along with learning fine inter-frame correspondence.

\paragraph{Siamese Natural Image Tracking} strategies leverage a Siamese architecture for matching between search and target templates, where the extracted spatial search and template features are matched via feature fusion or a similar matching module \cite{li2018high, li2019evolution, fan2019siamese, zhu2018distractor, fan2021cract, yu2020deformable, zhang2021learn}. With the rise of transformers, Siamese trackers have been extended to incorporate transformer-based models like Stark \cite{yan2021learning}, Mixformer \cite{cui2022mixformer}, among other methods cited in \cite{kugarajeevan2023transformers, wang2021transformer, chen2021transformer}.

\paragraph{Historical-Trajectory-based Natural Image Tracking} approaches leverage prompt-based methods to integrate relevant information. In particular, the temporal information is passed into the network as prompts to incorporate the historical trajectory information. ARTrack \cite{wei2023autoregressive} employs a decoder that receives these encodings as well as coordinates of the searched object from previous frames as spatio-temporal prompts for a trajectory proposal.
Another approach, SwinTrack \cite{lin2022swintrack}, uses a multi-head cross-attention decoder that leverages both the encoder output and a motion token, which represents the past object trajectory given previous bounding box predictions.

\paragraph{Device tracking in X-Ray:} Specifically for the tracking of devices in X-Ray images, multiple approaches have been proposed, including multiple Siamese-based architectures similar to those in natural image object tracking \cite{bromley1993signature,li2018high}.
Other methods such as Cycle Ynet \cite{lin2020cycle} employ a semi-supervised approach to address the lack of annotated frames in the medical domain, or leverage deep learning-based Bayesian filtering for catheter tip tracking \cite{ma2020dynamic}. One of the most recent approaches, ConTrack \cite{demoustier2023contrack}, uses a Siamese architecture and a transformer-based feature fusion model. To further refine the tracking, it incorporates a RAFT \cite{teed2020raft} model applied to catheter body masks for estimating optical flow.

\section{Methods}
We propose a novel Frame Interpolation Masked Autoencoder (FIMAE) approach to train a transformer model to extract spatio-temporal features based on a large internal dataset $\mathcal{D}_u$. 
The model is designed specifically to learn inter-frame correspondences over a large number of frames. The pretrained encoder is then used as backbone for the downstream tracking task using supervised learning on a dataset $\mathcal{D}_l$ (with expert annotations). The pretraining method and the tracking pipeline are explained in the following subsections.

\subsection{Self-supervised Model Training}

\paragraph{Learning space-time embeddings} Given the unlabeled dataset $\mathcal{D}_u$, $n$ frames are sampled from an arbitrary sequence $S_k \in \mathcal{D}_u$, $\forall k > 0$, where $S_{k,n} = [I_1, I_2, ..., I_n]$. All image frames are randomly cropped to a size of $(h,w) = 384 \times 384$ pixels on a sequence level (i.e., same crop applied to each image). Each input of size $(h,w)$ is spatially encoded into $n \times \frac{h}{16} \times \frac{w}{16}$ tokens of dimension $D_m$ with no temporal downsampling. 

\paragraph{Masking strategy based on frame interpolation} 
In order to learn features that capture fine spatial information and fine temporal correspondences between frames, we propose a novel masking strategy based on frame interpolation, that overcomes the limitation of the symmetrical tube masking proposed by VideoMAE~\cite{tong2022videomae}. Recall that the VideoMAE approach is limited in capturing fine inter-frame correspondences. Traditionally, in the domain of natural imaging, the frame interpolation task \cite{Jiang_2018_CVPR, Niklaus_2017_ICCV} is defined as the as a sum of forward warping and backward warping of any two neighboring frames (indexed by $t > 0$):


\begin{figure*}[t]
	\centering
	\includegraphics[width=1\columnwidth]{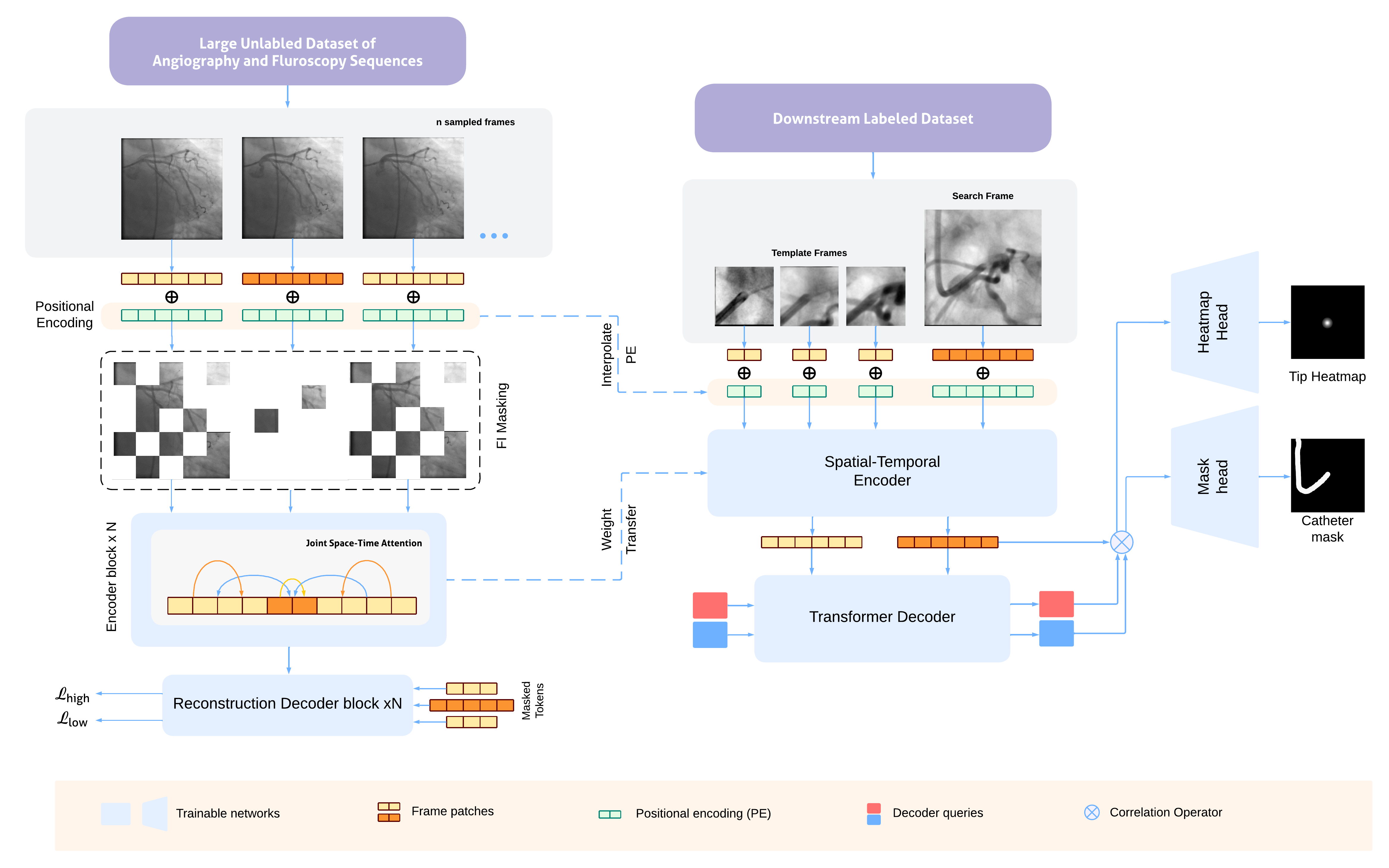} 
	\caption[Overview of our framework: Pretraining using \textbf{Frame Interpolation Masked Autoencoder (FIMAE)} (left) and downstream tracking model (right).]{Overview of our framework. First, the encoder is trained to learn spatio-temporal features from a large unlabeled dataset of angiography and fluorscopy using \textbf{Frame Interpolation Masked Autoencoder (FIMAE)} (left). Then, the weights are transfered into ViT encoder for feature extraction and feature matching for tracking the catheter tip (right). }
	\label{fig:workflow}
\end{figure*}

\begin{equation}
    I_{t+1} = \tau_{\theta_1}(I_t) + \tau_{\theta_2}(I_{t+2}) \enspace ,
\end{equation}
where $\tau_{\theta_1}$ denotes the forward warping operator and $\tau_{\theta_2}$ denotes the backward warping operator (parametrized by $\theta_1, \theta_2$). However, the change of appearance in coronary vessel structures in presence of contrast can be much more complex than natural images. Hence, a linear operation of forward and backward warping can limit the potential of the network. In our case, we reformulate this to a learning problem, seeking to optimize the parameters $\theta$ of a deep neural network to learn a combined warping operation $F$ as:

\begin{equation}
    I_{t+1} = F_{\theta}(I_t, I_{t+2}) \enspace .
\end{equation}

\begin{figure}
\centering
\includegraphics[width=.75\columnwidth]{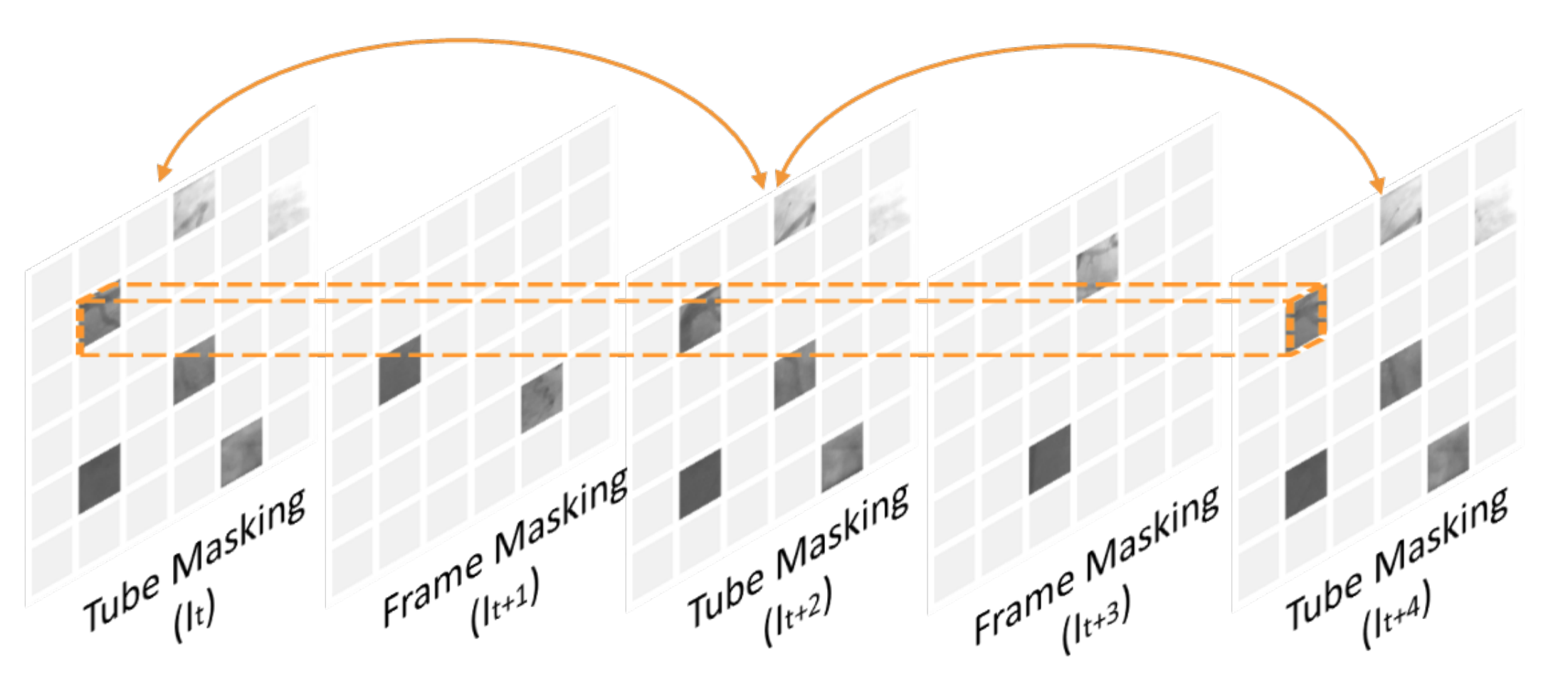}
\caption{Schematic visualization of tube-frame masking.}
\label{fig:tubeframemasking}
\end{figure}

In our approach, we use tube masking for every alternate frame with a ratio of $75\%$ and combine it with frame masking. However, with such high tube masking ratio, further masking an entire intermediate frame for frame interpolation can make the task extremely challenging. In addition, masking an entire frame may also lead the network to never attend to certain patch positions during training. Hence, we mask the intermediate frame randomly to a high ratio of $98\%$, instead. See Fig.~\ref{fig:tubeframemasking} for a schematic visualization.

Let $p_t \in \Omega_{tube}$ be the token indices of the tube masked tokens for frame $t$, where $\Omega_{tube}$ denotes the set of all tube masked token indices. Similarly, $q_t \in \Omega_{frame}$ refers to the frame masked token indices for frame $t$ in all randomly frame masked token indices. Mathematically, if $\rho$ is the probability for masking, $\Omega_{tube} \sim Bernoulli(\rho_{tube})$ where different time $t$ shares the same value. On the other hand, $\Omega_{frame} \sim Bernoulli(\rho_{frame})$, and is drawn uniquely for each frame at $t$.
Let $p'_t \in \Omega'_{tube}$ and $q'_t \in \Omega'_{frame}$ be the sets of remaining visible token indices. Combining tube and frame masking strategies, we obtain the following reconstruction objective for any $3$ given frames: 

\begin{equation}
    I_t, I_{t+1}, I_{t+2} = F_{\theta}(I_t(p'_t), I_{t+1}(q'_{t+1}), I_{t+2}(p'_{t+2}))
\label{eq:3frames}
\end{equation}
where $0 < t < n-1$ denotes the index of an arbitrary frame from the sampled sequence, $I_t(p'_t)$ denotes the visible patches of frame $I_t$ with tube/frame masking applied. The 3-frame objective shown in Eq.~\ref{eq:3frames} can be generalized to all $n$ frames.





\paragraph{Encoder-Decoder Training} The unmasked patches are passed through a ViT encoder which adopts a joint space-time attention. That is, each token for frame $t$, is projected and flattened into $D_m$-dimensional vector query, key and value embedding: $(q_t, k_t, v_t)$. The joint space-time attention is based on the concatenated vectors:

\begin{equation}
	\label{eqn:attention}
	\operatorname{Attention}(\mathrm{Q}, \mathrm{K}, \mathrm{V}) =\operatorname{softmax}\left(\frac{Q K^T}{\sqrt{d}}\right) V,
\end{equation}

where the variables $(Q, K, V)$ are defined as $\mathrm{Q}=\operatorname{Concat}(q_1, q_2, \ldots, q_n), \\
\mathrm{K} = \operatorname{Concat}(k_1, k_2, \ldots, k_n), \mathrm{V}= \operatorname{Concat} (v_1, v_2, \ldots, v_n)$, for $n$ sampled consecutive frames. 
The encoded visible patches are then concatenated with learnable masked tokens. A lightweight transformer decoder attends to the encoded patches and the masked token to reconstruct the initially masked patches. The decoder incorporates additional positional encoding to ensure the correct positions of the masked and unmasked patches as per the original frames. 

\paragraph{Pretraining Loss Function} We use a weighted mean squared error (MSE) loss, $\mathcal{L} = \mathcal{L}_{\text{tube}} + \gamma \mathcal{L}_{\text{frame}}$ between the masked tokens and the reconstructed ones in the pixel space based on the masking strategy, where $\gamma$ is the weighting factor:

\begin{equation}
	\mathcal{L}_{\text{tube}}=\frac{1}{|\Omega_{tube}|}\sum_{t={2\eta + 1}}^n \sum_{p_t \in \Omega_{tube}}\|I_t(p_t)-\hat{I}_t(p_t)\|^2,
\end{equation}

\begin{equation}
	\mathcal{L}_{\text{frame}} = \frac{1}{|\Omega_{frame}|}\sum_{t={2\eta+2}}^n \sum_{q_t \in \Omega_{frame}}\|I_t(q_t)-\hat{I}_t(q_t)\|^2,
\end{equation}

where $I$ is the input image, $\hat{I}$ is the reconstructed image, and $0 \leq \eta \leq (n-2)/2 $.
We use a weighted loss for reconstruction to compensate for the imbalance between low masked frames (less reconstruction tokens) and highly masked frames (more reconstruction tokens). The variable $\gamma$ is defined as the ratio of number of $\Omega_{tube}$ tokens and the number of $\Omega_{frame}$ tokens. 

\subsection{Downstream Application: Device Tracking}
Particularly for tracking the tip of the catheter, our goal is to track its location, $\hat{y}_t = (u_t, v_t)$ at any time $t, t>0$ given a sequence of X-ray images $\{ I_t\}_{t=1}^n$  with a known initial location of the catheter tip $y_1 = (u_1, v_1)$ on the labeled dataset $\mathcal{D}_l$. We consider the sequences $S_k \in \mathcal{D}_l$, $\forall k>0$ to have only few annotated labels, $S_{k,n} = [(I_1, y_1), (I_2), ... , (I_7, y_7), (I_{8}), ...]$. To identify the location of the tip of the catheter at current search frame, existing approaches build a correlation with a template frame. The template frame is usually a small crop around the catheter tip location from a previously predicted frame. Similar to ConTrack, during training we use three template frames that are cropped from the first annotated frame and the previous two annotated frames, respectively. We use the current frame for template frames if no previously annotated frames are available.
During inference, the initial location of the catheter tip serves as the first template crop and is kept intact. The rest two template frames are updated dynamically based on the model's predictions.

\paragraph{Feature transfer} The spatio-temporal transformer backbone inputs three template frames and a search frame as four distinct frames. We interpolate the positional encoding from the pretraining frame positions appropriately to ensure that the network distinguishes each template and search frame as distinct frames. In particular, each template frame and the search frame correspond to the positions of center crops of individual frames in the pretraining setup. Therefore, the encoder inputs $\operatorname{Concat}(te_1, te_2, te_3, se)$, where $te_{1,2,3}$ and $se$ are template patches and search patches respectively. Given that transformers are isotropic models, we obtain an encoded feature set, $f_c = \operatorname{Concat}(f_{te_1}, f_{te_2}, f_{te_3}, f_{se})$.
The spatio-temporal transformer backbone is trained to extract fine inter-frame correspondences. Hence, this results in a joint feature extraction and feature matching between the template frames and the search frame. The overview of the proposed model is depicted in Fig. \ref{fig:workflow}.

\paragraph{Multi-task Transformer Decoder} We use a lightweight Transformer decoder similar to the original Transformer model \cite{vaswani2017attention}. First, all the features $f_c$ are projected to a lower dimension $d_m$. The decoder uses two learnable query tokens $(h_d, m_d)$, each for a heatmap head and a mask head. Then, each layer first computes attention on the query tokens as per Eq. \ref{eqn:attention}. It is followed by cross-attention with encoded features $f_c$, where key and value embeddings are computed by projecting the features $f_c$ to dimension $d_m$. The resulting query tokens are then correlated with the search features, unflattened and passed through a CNN head:

\begin{equation}
	P_h = \operatorname{Conv_h}(\operatorname{Unflatten}(\operatorname{corr}(f_{se}, h_d))),
\end{equation}      

\begin{equation}
	P_m = \operatorname{Conv_m}(\operatorname{Unflatten}(\operatorname{corr}(f_{se}, m_d))).
\end{equation}  

The final tip coordinates are obtained by $\hat{y} = \operatorname{max}(P_h)$, where $P_h$ and $P_m$ refer to predicted heatmap of the catheter tip and predicted mask of the catheter respectively. We compute soft dice loss $\mathbf{L}_{\text {dice }} = \mathbf{L}_h + \lambda \mathbf{L}_m$, for both heatmap and mask predictions, given by:

\begin{equation}
	\mathbf{L}_h=\frac{2 * \sum G_h * P_h}{\sum G_h^2+\sum P_h^2+\epsilon},
\end{equation}  

\begin{equation}
	\mathbf{L}_m=
	\begin{cases}
		\frac{2 * \sum G_m * P_m}{\sum G_m^2+\sum P_m^2+\epsilon}, & \text{if }  G_m \text{ exists} \\
		0 & \text{otherwise}
	\end{cases},
\end{equation}

where $G$ represents ground truth labels, and $\lambda$ is the weight for weighting mask loss.

\section{Experiments and Results}
\subsection{Dataset}
An unlabeled internal dataset $\mathcal{D}_u$ of coronary X-ray sequences is utilized to pretrain our model. $\mathcal{D}_u$ consists of 241,362 sequences collected from 21,589 patients, comprising 16,342,992 frames in total. It contains both fluoroscopy ("Fluoro") and angiography ("Angio") sequences.
We randomly sample 10 frames at a time, with varying temporal gaps between them, ranging from 1 to 4 frames. We repeat the last frame in sequences where the number of frames is less than 10.
The model is then pretrained for 200 epochs with a learning rate of $1\mathrm{e}{-4}$. 

\begin{figure}[ht]
	\centering
    \includegraphics[width=0.75\linewidth ]{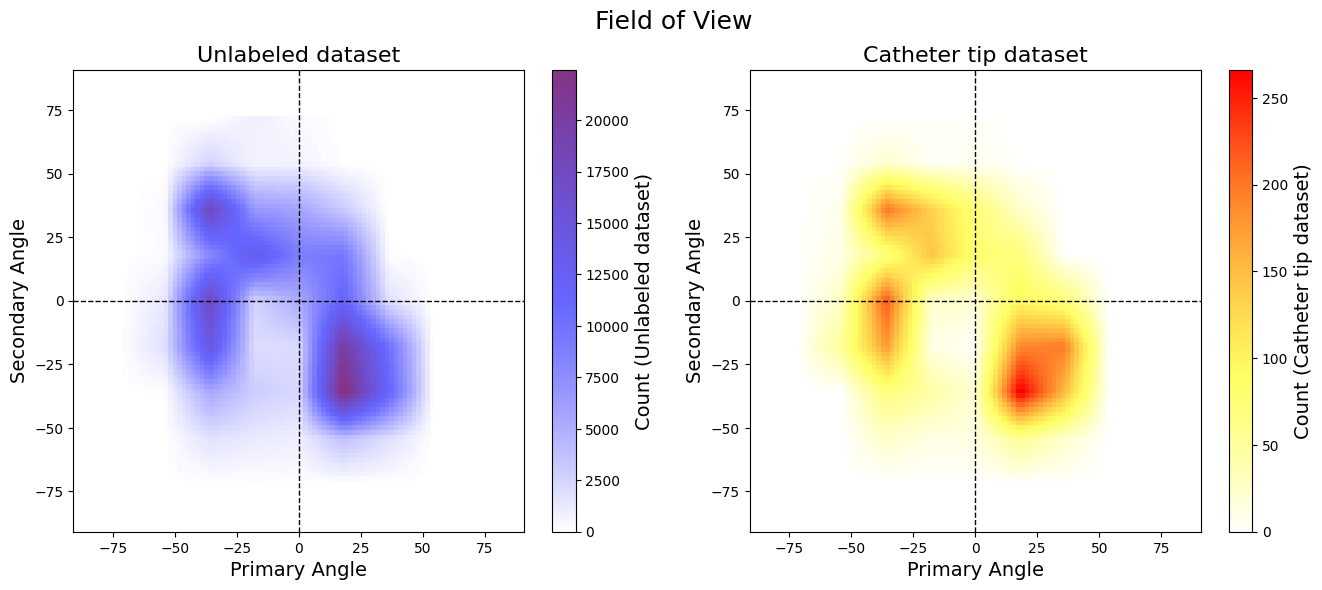}
	\caption[Distribution of the datasets based on the Field of View]{Distribution of the datasets based on the Field of View (Positioner Primary angle and Positioner Secondary angle): The left plot denotes the unlabled dataset ($\mathcal{D}_u$) and the right plot denotes the catheter tip dataset ($\mathcal{D}_l$).}
	\label{fig:FOV}
\end{figure}

For the downstream tracking task, we use dataset $\mathcal{D}_l$.
Note that $\mathcal{D}_l \cap \mathcal{D}_u = \emptyset$. The distribution of field of view for both $\mathcal{D}_u$ and $\mathcal{D}_l$ is depicted in Fig. \ref{fig:FOV} and is estimated based on the Positioner angles. The Positioner Primary Angle is defined in the transaxial plane at the imaging device's isocenter with zero degrees in the direction perpendicular to the patient's chest and +90 degrees at the Patient left hand side and -90 at the Patient right hand side. The Positioner Secondary Angle is defined in the Sagittal Plane at the imaging device's isocenter with zero degrees in the direction perpendicular to the patient's chest. Fig.~\ref{fig:FOV} shows that the distribution of the sequences in both the datasets are concentrated around similar Positioner angles. Other attributes from both the datasets $\mathcal{D}_l$ and $\mathcal{D}_u$ are depicted in Table \ref{tab:datastats}. 

\begin{table}[ht]
\centering
\caption{Dataset Statistics (Range and median) for Unlabeled Dataset ($\mathcal{D}_u$) and Catheter Tip Dataset ($\mathcal{D}_l$)}
\label{tab:datastats}
\begin{tabular}{ccccc}
Attributes              & \multicolumn{2}{c}{Unlabeled Dataset ($\mathcal{D}_u$)} & \multicolumn{2}{c}{Catheter Tip Dataset ($\mathcal{D}_l$)} \\ \hline
                        & Range                 & median        & Range                  & median          \\ \hline
FPS                     & 1 to 30               & 15            & 1 to 30                & 15              \\
\#Frames                & 1 to 552              & 83            & 4 to 920               & 77              \\
Resolution (mm/pixel)   & 0.129 to 0.616        & 0.279         & 0.108 to 0.368         & 0.279           \\
Peak Kilo Volt          & 45.16 to 125.0        & 87.1          & 61.0 to 125.0          & 86.3            \\
Tube Current (mA)       & 1.0 to 928.0          & 757.0         & 7.0 to 904.0           & 740.0           \\
Exposure Time (msec)    & 3 to 20235            & 522           & 5 to 14160             & 503             \\ \hline
\end{tabular}
\end{table}

\begin{table}[ht]
\centering
\caption{SNR of different categories in catheter tip dataset ($\mathcal{D}_l$)}
\label{tab:snr}
\begin{tabular}{ccc}
Fluoro & Angio & Devices \\ \hline
24.72 dB & 21.38 dB & 23.64 dB   \\ \hline  
\end{tabular}
\end{table}

\begin{figure}[ht]
	\centering
    \includegraphics[width=1\linewidth ]{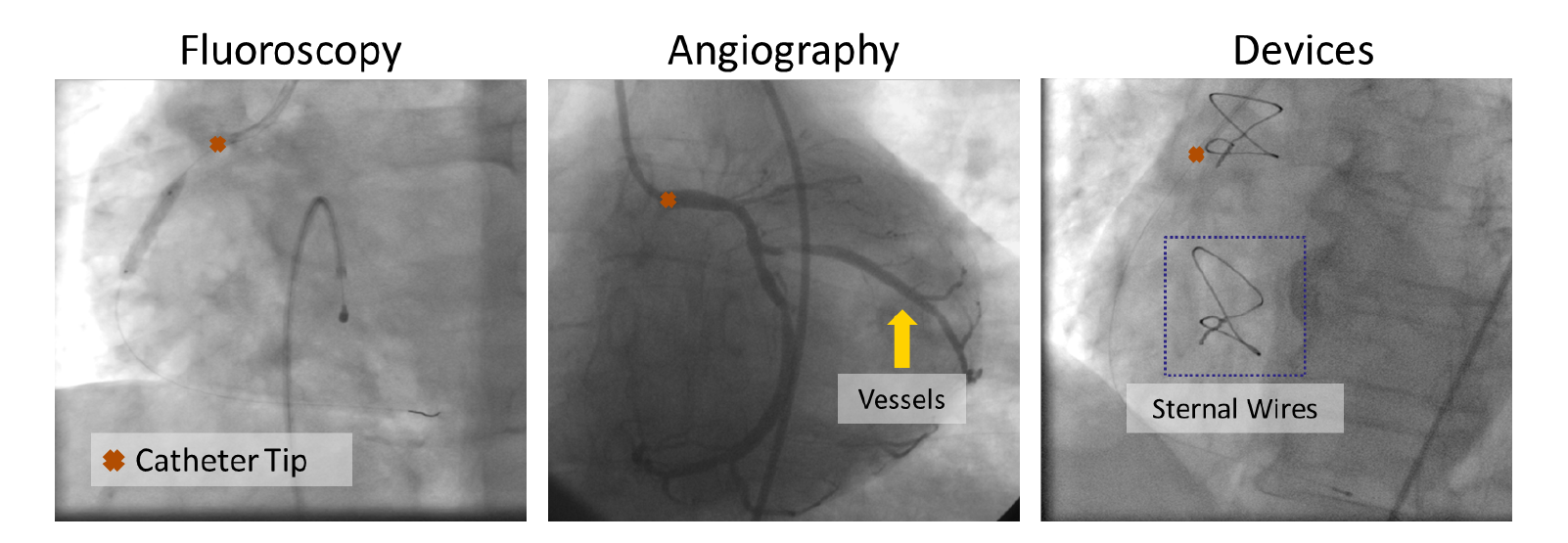}
	\caption{Visualization of tip of the catheter in fluoroscopy, angiography and cases with other devices.}
	\label{fig:example_cases}
\end{figure}

The annotations on the frames in $\mathcal{D}_l$ represent the coordinates of the tip of the catheter, which are converted to Gaussian heatmaps with standard deviation of $\approx$ 5mm.  Mask annotations of the catheter body are also available for a subset of these annotated frames. 
In an average, the catheter body takes up 0.009\% of the total area of a frame.
The training and validation set consists of 2,314 sequences totaling 198,993 frames, out of which 44,957 have annotations. 
In this set, 2,098 sequences are Angio and only 216 sequences are Fluoro.
The test set consists of 219 sequences, where all 17,988 frames are annotated. For evaluation, we split the test set into three categories: 94 Fluoro sequences (8,494 frames, 82 patients), 101 Angio sequences (6,904 frames, 81 patients), and 24 Devices sequences (2,593 frames, 10 patients) \cite{demoustier2023contrack}. The latter category, "Devices", covers all sequences where sternal wires are present, which cause occlusion, thus further increasing the difficulty of catheter tip tracking. Examples of these cases are illustrated in Fig. \ref{fig:example_cases}.
The SNR of the image intensity at the catheter tip with respect to the background is shown in Table~\ref{tab:snr}, further quantifying the challenge of tracking. The SNR was calculated based on the following formula: 

\begin{equation}
    SNR = 20\log_{10}\frac{P_w}{\sigma_f}
\end{equation}
where $P_w$ is the mean intensity in the window of $6 \times 6$ ($\approx$ 2mm $\times$ 2mm) and $\sigma_f$ denotes the standard deviation of the intensity of the background in the window of $30 \times 30$ ($\approx$ 10mm $\times$ 10mm) with the catheter tip as the centre of both windows.

We follow the same image pre-processing pipeline as ConTrack, i.e., we resample and pad to size of 512 × 512 with 0.308 mm isotropic pixel spacing. We use 160$\times$160 crops for search image, and 64$\times$64 crops for template images. We train our model for 100 epochs, with a learning rate of $2\mathrm{e}{-4}$ using AdamW optimizer and Cosine Annealing scheduler with warm restarts. 

\subsection{Performance Evaluation} 

\begin{table*}[ht]
	\centering
	\caption[Comparison study of sequence-level tracking errors and runtime for different methods.]{Comparison study of sequence-level tracking errors (mean euclidean distance) and runtime for different methods for catheter tip tracking in coronary X-ray sequences. Best numbers are marked in bold black and the second best is marked in blue. We also show the performance of different versions of ConTrack. ConTrack-base refers to its base version which has no additional modules, ConTrack-mtmt refers to multi-task and multi-template version and  ConTrack-optim is its final optimal version which has all modules including flow refinement. }
	\begin{tabular}{c|c|c|c|c|c|c}

		Models & Median & Mean & Std & 95 Percentile & Max & Speed\\ 
		& (mm) $\downarrow$ & (mm) $\downarrow$ & (mm) $\downarrow$ & (mm) $\downarrow$ & (mm) $\downarrow$ & (fps) $\uparrow$ \\ \hline
		SiameseRPN \cite{li2018high}& 7.13 & 9.01 & 6.81 & 22.37 & 46.23 & 18 \\ 
		STARK \cite{yan2021learning}& 2.65 & 4.14 & 4.93 & 9.24 & 31.34 & 22\\ 
		MixFormer \cite{cui2022mixformer} & 2.68 & 5.15 & 7.1 & 19.20 & 49.29 & 20 \\ 
		Cycle Ynet \cite{lin2020cycle} & 1.96 & 2.68 & 2.4 & 6.75 & 21.04 & \textbf{109} \\
        ConTrack-base \cite{demoustier2023contrack} & 1.13 & 2.17 & 3.75 & 6.34 & 31.35 & 21 \\
        ConTrack-mtmt \cite{demoustier2023contrack} & 1.12 & 1.97 & 3.61 & 5.53 & 30.37 & 19 \\
		ConTrack-optim \cite{demoustier2023contrack} & \textcolor{blue}{1.08} & \textcolor{blue}{1.63} & \textcolor{blue}{1.7} & \textcolor{blue}{5.18} & \textcolor{blue}{13.32} & 12 \\ 
		Ours & \textbf{1.02} & \textbf{1.44} & \textbf{1.35} & \textbf{3.52} & \textbf{10.23} & \textcolor{blue}{42}\\ 
        \hline
	\end{tabular}
	\label{table:main}
\end{table*}

We evaluate our work against state-of-the-art methods, explore the impact of the proposed pretraining strategy, and investigate whether complex additional tracking refinement modules are necessary. All the evaluations are performed based on expert annotations.

\begin{figure}[ht]
	\centering
	\includegraphics[width=\columnwidth]{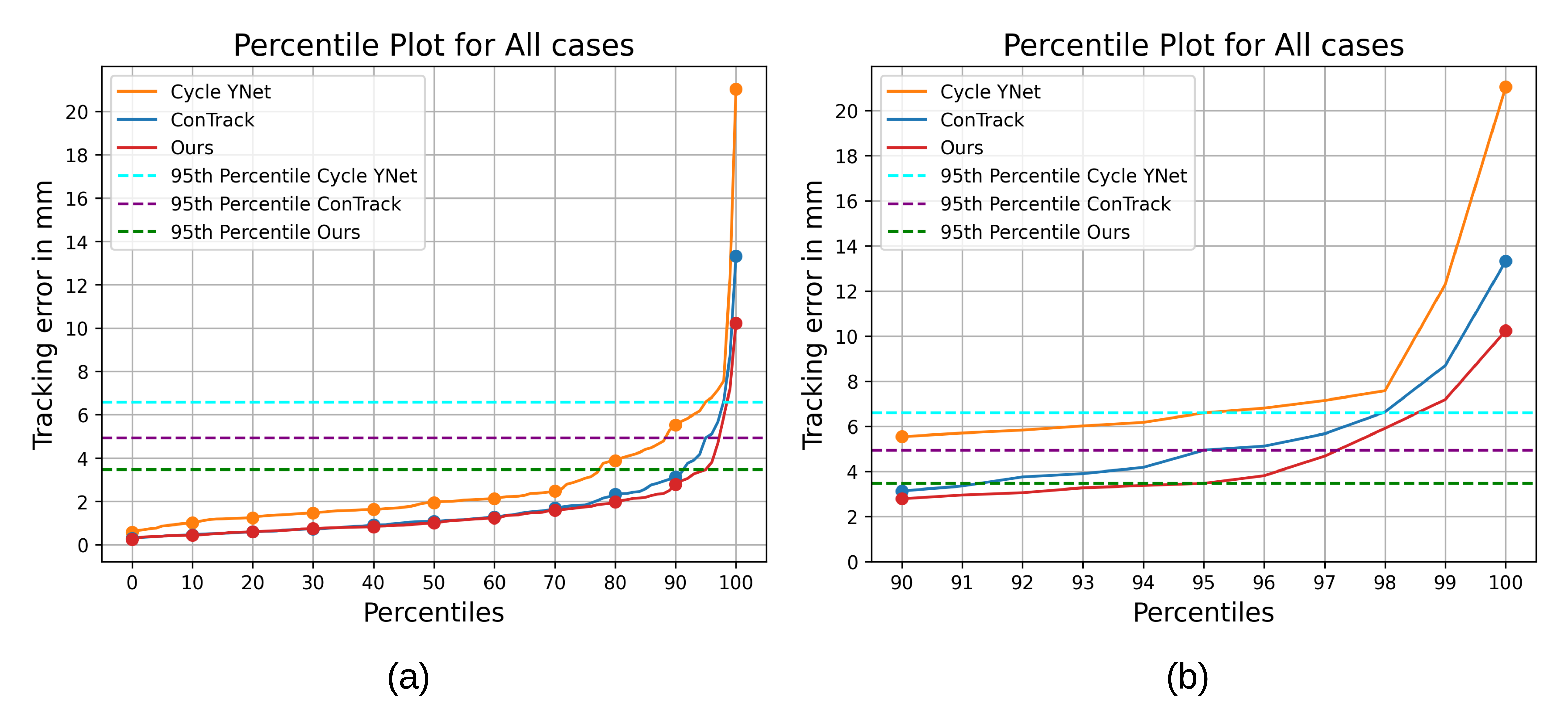}
	
	\caption[Percentile plot of Cycle YNet, ConTrack and ours (a) for all test cases and (b) zoomed in for percentiles from 90th to 100th.]{Percentile plot of Cycle YNet, ConTrack and ours (a) for all test cases and (b) zoomed in for percentiles from 90th to 100th. 95th percentile of our method's performance is lesser than the average diameter of the vessels ($\approx$ 4mm).}
	\label{fig:percplot}
\end{figure}

\paragraph{Benchmarking against State-of-the-Art} We report the performance of our model against the state of the art device tracking models in Table~\ref{table:main}. Here, we evaluate the euclidean distance error in mm between the prediction and the ground truth annotations. 
Overall, our method demonstrates the best performance on the test dataset, excelling in both precision and robustness. Our approach significantly reduces the overall maximum error, e.g., by 66.31\% against the comparable version of ConTrack (ConTrack-mtmt) and by 23.20\% against ConTrack-optim, a highly optimized solution leveraging multi-stage feature fusion, multi-task learning and flow regularization. In comparison to previous state-of-the-art approaches, our approach results in fewer failures, as depicted by the error distribution in Fig. \ref{fig:percplot}. Atleast 95\% of the all test cases has an error below the average diameter of the vessels ($\approx$ 4mm).
Notably, our approach stands out from other tracking models by eliminating the need for a two-stage process involving the extraction of spatial features and subsequent matching using feature fusion. Instead, our spatio-temporal encoder jointly performs both. 

Other approaches often require two or more forward passes for the two-stage processing to incorporate varying template-search size, which increases computational complexity. This is further amplified by the inclusion of additional modules, such as multi-task decoders and the flow-refinement network in ConTrack-optim \cite{demoustier2023contrack}. In contrast, our model accomplishes the task with a single forward pass for both the multiple templates and the search frame. The only additional modules in our model are the two CNN heads for multi-task decoding. This design choice enables us to achieve a significantly higher real-time inference speed of 42 fps on a single Tesla V100 GPU without compromising on accuracy, as shown in Fig.~ \ref{fig:err_speed}. Despite Cycle Ynet \cite{lin2020cycle} also relying on multiple forward passes for feature extraction, its simplicity and computationally friendly CNN architecture allows it to reach higher speed, albeit at the expense of accuracy and robustness.

\begin{table}[!ht]
	\centering
	
	\caption[Study of effect of pretraining startegies on the performance of the catheter tip tracking.]{Study of effect of pretraining startegies on the performance of the catheter tip tracking. Pretraining is performed either on our internal dataset (denoted as $\mathcal{D}_u$) or on natural images (in case of the first approach).}
	\begin{tabular}{c|c|c|c|c}

		Pretraining & Median & Mean & Std & Max \\ 
		strategy & (mm) & (mm) & (mm) & (mm) \\ \hline
		VideoMAE-Kinetics & 1.93 & 3.67 & 4.95 & 36.99 \\ 
		  VideoMAE ($\mathcal{D}_u$) & 1.48 & 2.75 & 4.64 & 53.26 \\ 
		SiamMAE ($\mathcal{D}_u$) & 1.54 & 2.79 & 3.44 & 23.76 \\ 
		Ours: FIMAE ($\mathcal{D}_u$) & \textbf{1.02} & \textbf{1.44} & \textbf{1.35} & \textbf{10.23} \\ 
		\hline
	\end{tabular}
    \label{table:pretraining}
\end{table}

\begin{figure*}[t]
	\centering
    \includegraphics[width=1\linewidth ]{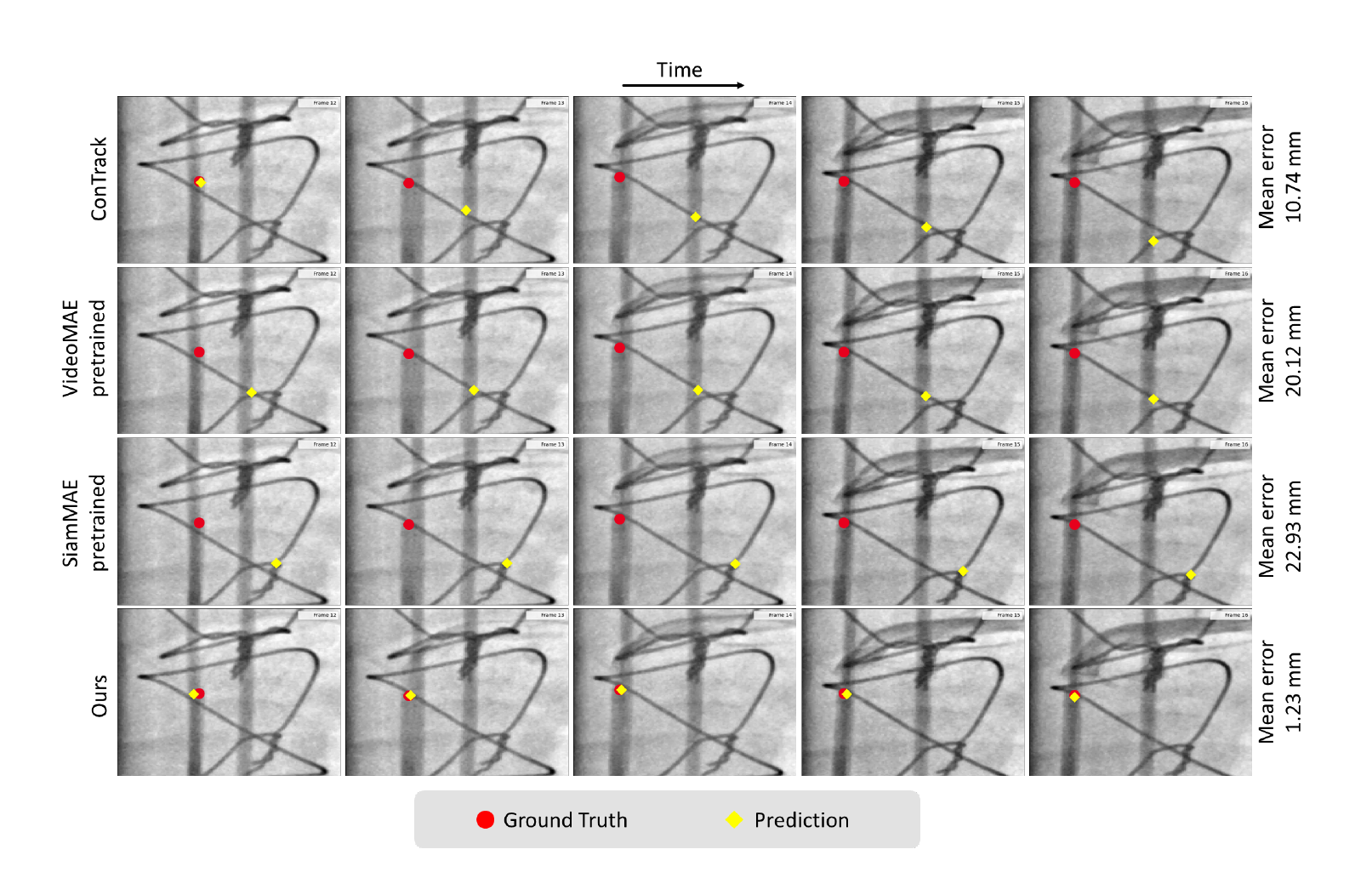}
	\caption[\textbf{Qualitative Results.} Comparison of different methods on a challenging sequence of angiography, where tracking receives obstruction from vessels and sternal wires (other devices)]{\textbf{Qualitative Results.} Comparison of different methods on a challenging sequence of angiography, where tracking receives obstruction from vessels and sternal wires (other devices). Note that the images have been cropped around the region of interest for better visualization. The mean error depicted in the figure is the average error computed over the entire sequence.}
	\label{fig:qual_1}
\end{figure*}

\paragraph{Impact of Pretraining} Next, we focused on the impact of pretraining by comparing tracking performance utilizing our proposed pretraining strategy (FIMAE) against current prevalent pretraining methods for sequential image processing, see Table \ref{table:pretraining}. 
The findings indicate that pretraining on domain-specific data, as opposed to natural images (VideoMAE-Kinetics), offers significant advantages. However, even when including the models trained on $D_u$ (VideoMAE and SiamMAE) into the comparison, our model surpasses all by more than 30\% across all reported metrics. VideoMAE lacks fine temporal correspondence between frames, leading to non-efficient feature matching between template and search frames. While SiamMAE has the ability to learn inter-frame correspondence, it relies on only two frames at a time, which is insufficient to fully capture the underlying motion. Qualitative results are shown in Fig. \ref{fig:qual_1}, based on a challenging angiography sequence with contrast-based device obstruction and other visible sternal wires. The figure shows how our model is able to handle this challenging case by  not losing track of the tip of the catheter where the other models fail to differentiate the catheter from the sternal wires.

\begin{figure*}[ht]
	\centering
    \includegraphics[width=1\linewidth ]{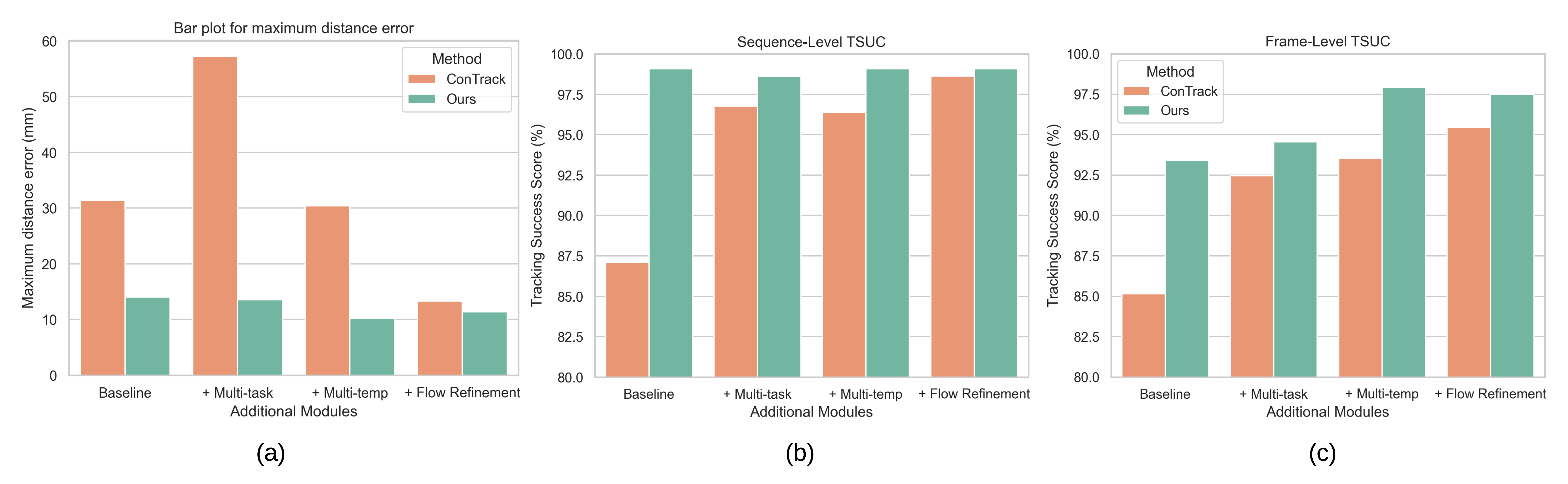}
    
	\caption{Comparison of robustness between our method and different versions of Contrack, via (a) maximum distance error ($\downarrow$), (b) sequence-level TSUC ($\uparrow$) and (c) frame-level TSUC ($\uparrow$).}
	\label{fig:bar_plots}
\end{figure*}

\paragraph{Performance without Complexity} The strength of our approach comes from the pretrained spatio-temporal features that facilitate effective feature matching between the template frames and the search frame. Another key advantage is its prior understanding of the inherent cardiac/respiratory motion. This knowledge significantly reduces or even eliminates the impact of additional modules such as flow refinement.  Our approach thereby achieves high robustness in tracking, with minimal variations across different additional modules such as multi-task.
To illustrate this, Fig. \ref{fig:bar_plots}(a) highlights the relative stability of the maximum error across different versions of our model compared to the high volatility observed in ConTrack under different module configurations. In addition, ConTrack reaches its best performance only when utilizing all modules, in particular including flow-refinement, which in turn leads to increased inference time. Contrary to ConTrack, adding the flow refinement module to our model even reduced its performance marginally in terms of accuracy ($1.54$ mm) and robustness (max error of $11.38$ mm). We postulate that this is attributable to the fact that while flow refinement can indeed learn intricate temporal correspondences between the previous and current frames, it can also propagate noise originating from inaccurately predicted catheter masks.

To further assess the robustness of the tracking systems, we introduce Tracking Success Score (TSUC), which draws parallels with most tracking benchmarks prevalent in single object tracking in the natural image domain \cite{fan2019lasot}. TSUC is computed as the ratio of number of instances (frame or sequence), in which the distance error falls below a specific threshold, to the total number of instances. To establish a relevant threshold, we set it at twice the average vessel diameter in our test dataset ($\approx 8mm$). Fig.~\ref{fig:bar_plots}(b) and Fig.~\ref{fig:bar_plots}(c) summarize the results for sequence-level and frame-level TSUC respectively. Our approach consistently achieves an impressive 99.08\%
sequence-level TSUC across all additional modules, with only a small drop to 98.61\% in the multi-task configuration. At the frame level, our optimal version (multi-task multi-template) yields a TSUC of 97.95\%, compared to 93.53\% for ConTrack under the same configuration. ConTrack achieves its best frame-level TSUC of 95.44\% using the flow-refinement variant.

\begin{figure}[ht]
	\centering
	\includegraphics[width=0.5\linewidth]{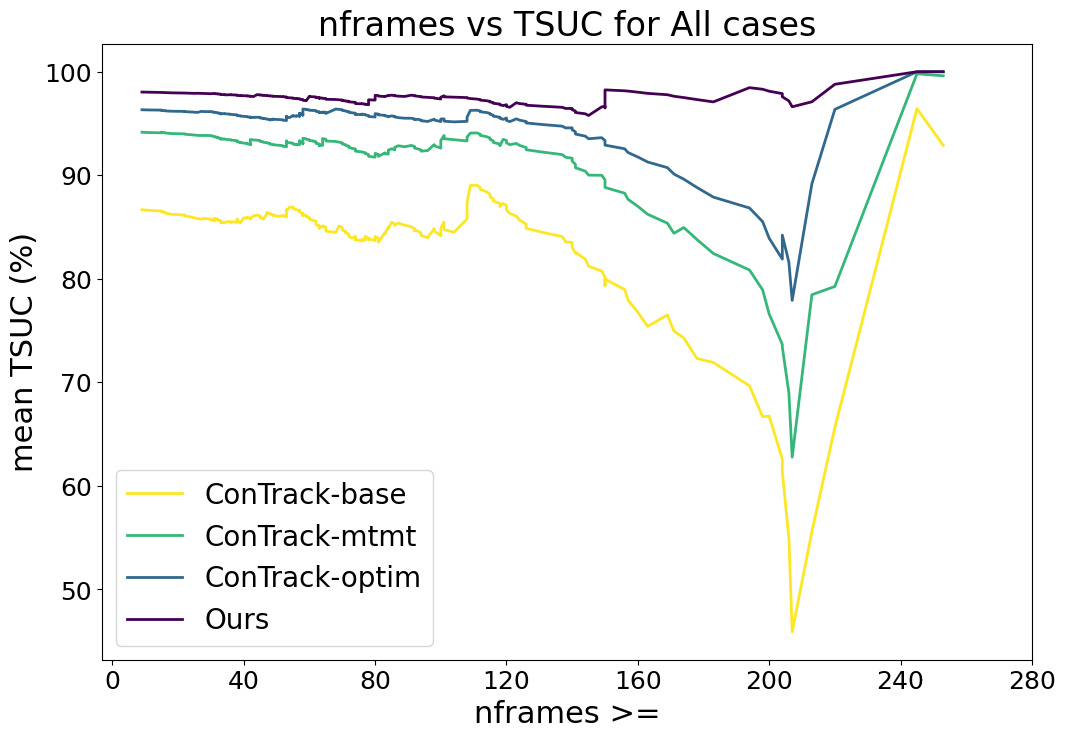}
	\caption[Robustness with respect to the sequence length.]{ Robustness with respect to the sequence length: mean TSUC for all sequences greater than the frame count (nframes). Note the dataset consists of only 4 sequences with frame count greater than 210.}
	\label{fig:tsuc_nframes}
\end{figure}

The robustness of a method is also influenced by its ability to effectively handle long sequences, as the accuracy of current frame predictions is dependent on previous frame predictions, resulting in gradual accumulation of errors over time. We examine the mean TSUC for sequences exceeding a certain frame count (nframes) in Fig.~\ref{fig:tsuc_nframes}. The plot shows that our method consistently demonstrates stable TSUC values across various sequence lengths, indicating its robust performance. Conversely, different versions of the ConTrack exhibit a gradual decline in mean TSUC as the frame count threshold increases, suggesting a reduced reliability in predicting outcomes over extended sequences.

\begin{figure}[ht]
	\centering
	\includegraphics[width=.5\columnwidth]{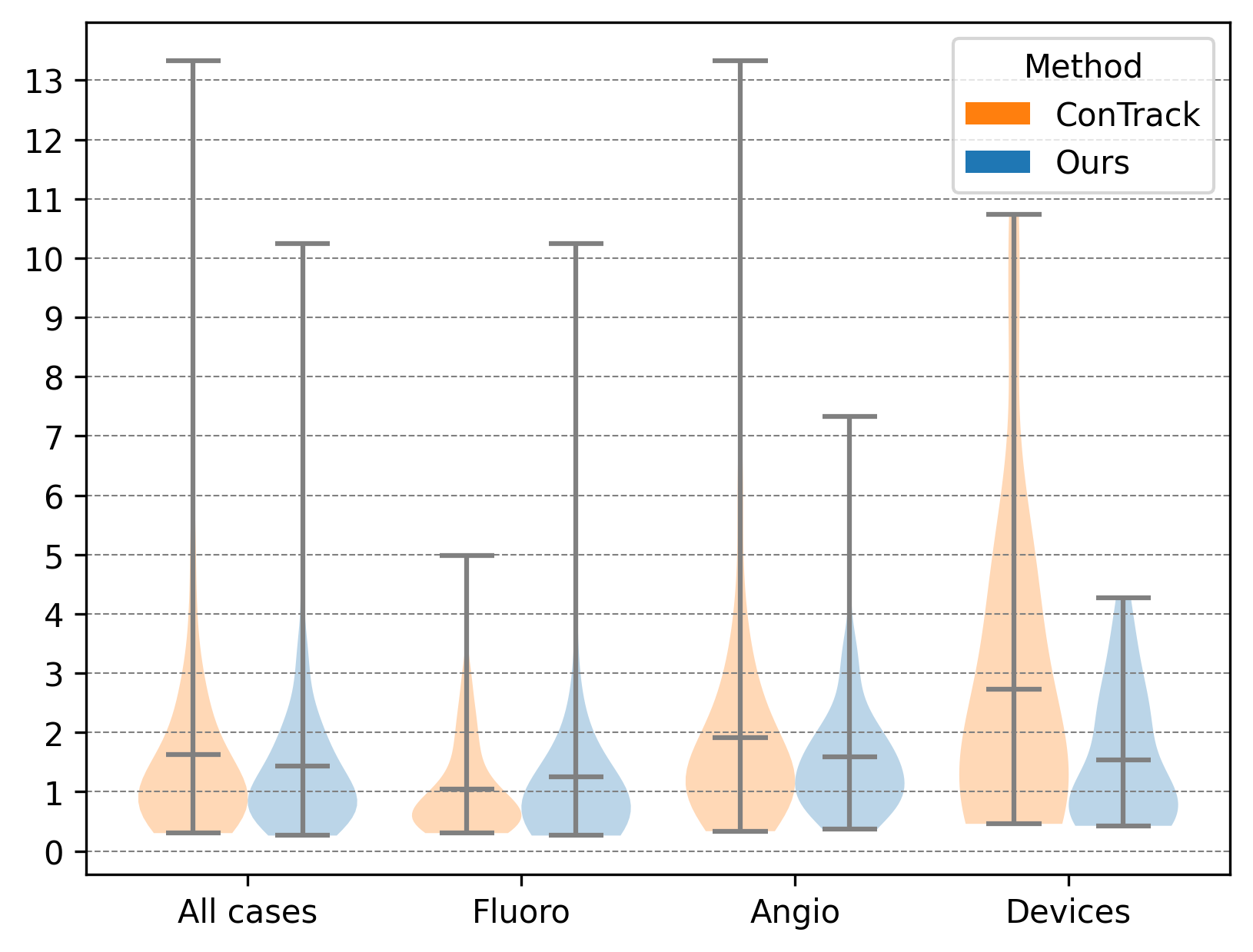} 
	\caption{Breakdown of different cases in a violin plot for comparison of our method with ConTrack.}
	\label{fig:violin_cases}
\end{figure}

\begin{figure}[!ht]
	\centering
	\includegraphics[width=1\columnwidth]{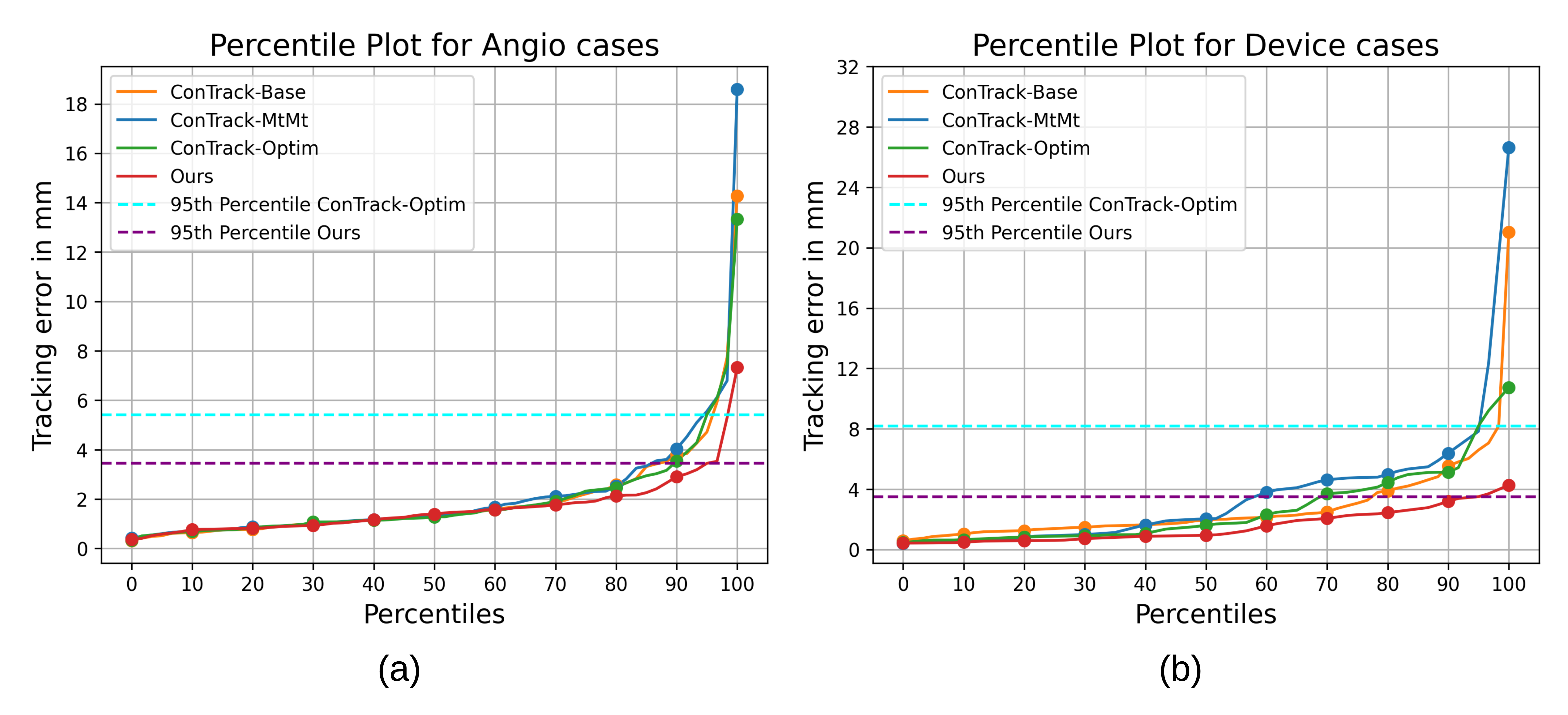}
	
    \caption{Percentile plots of different versions of ConTrack and Ours for (a) Angio Cases and (b) Device cases.}
	\label{fig:perplot}
\end{figure}

\begin{figure}[ht]
    \centering
    \includegraphics[width=\linewidth]{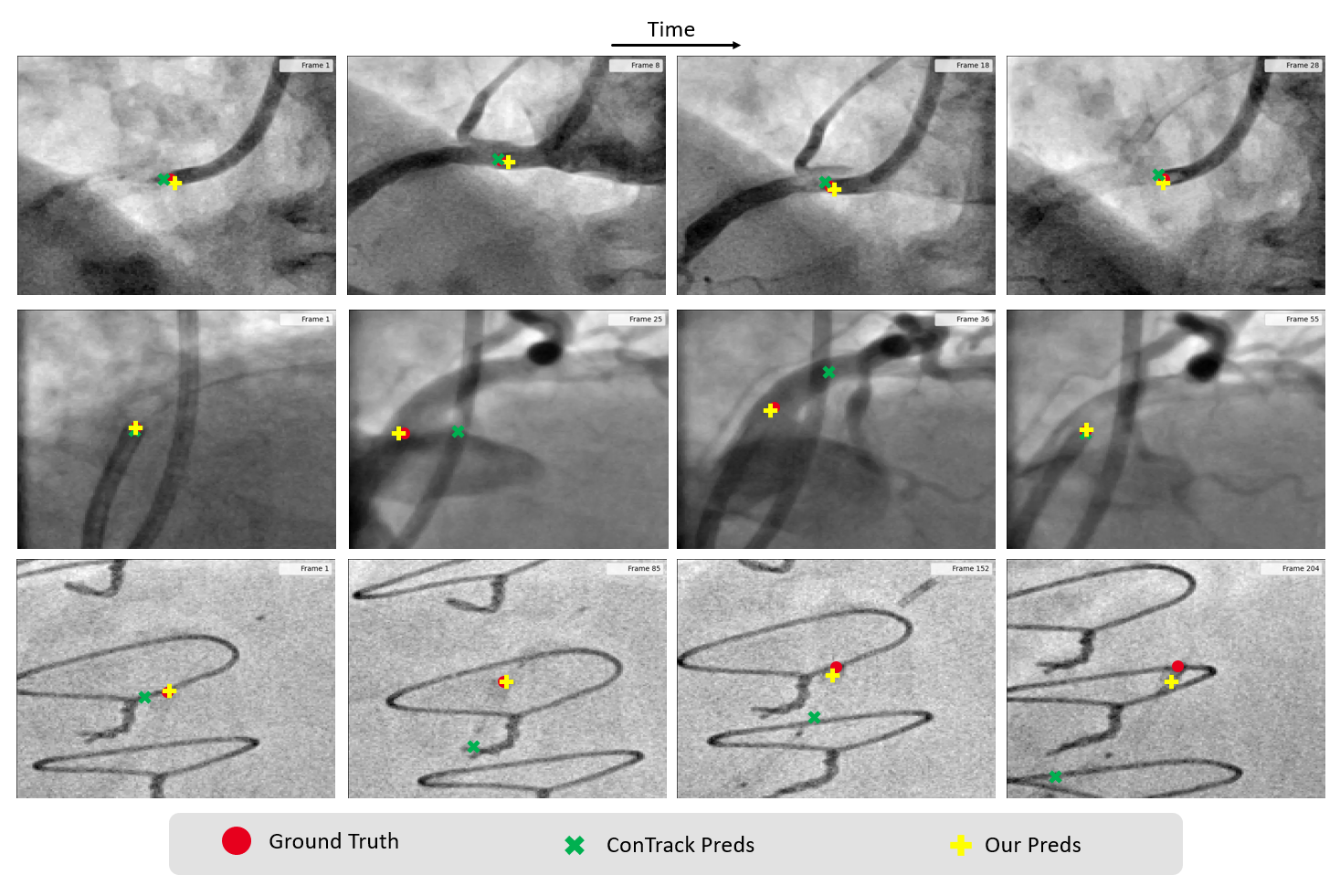}
    \caption[Visualization of predictions of ConTrack and our model in two Angio sequences (top two) and an extra device case (bottom).]{Visualization of predictions of ConTrack and our model in two Angio sequences (top two) and an extra device case (bottom). Note that the frames are sampled randomly from the sequence for visualization.}
    \label{fig:qual_more}
\end{figure}

\paragraph{Performance breakdown for different cases} We further conduct detailed comparison with the best-performing state-of-the-art method, ConTrack, for the different image categories defined earlier, see Fig.~\ref{fig:violin_cases}. We further compare our model's performance with ConTrack for the challenging cases, i.e., angiography and devices, via percentile plots in Fig. \ref{fig:perplot}. In the cases of angiography, our method shows 15\% improved accuracy and  45\% reduction in maximum error. Similarly for the devices (occlusion) category, where we achieve 43\% better accuracy and 60\% reduction in maximum error (Fig.~\ref{fig:violin_cases} and Fig. \ref{fig:perplot}). Our model's performance on Angio and Devices cases is compared qualitatively with ConTrack in Fig. \ref{fig:qual_more}. The example cases in the figure shows the effectiveness of our approach in the presence of complex occlusions from the vessels and sternal wires. ConTrack achieves a better performance than our method in Fluoro cases 
with a slightly better median and lesser maximum error. However, for Fluoro, ConTrack achieves a TSUC of 99.01\% (inaccurate in 1 sequence) compared to our model's TSUC of 97.69\% (inaccurate in 3 sequences). The inaccuracy of our model is seen in sequences where the visibility of the catheter is faint due to low-dose X-rays.
We hypothesise that this is due to Transformer's architecture that uses $16 \times 16$ non-overlapping patches making it less effective towards faint visibility in low-dose X-rays compared to CNNs in ConTrack, which uses overlapping $3 \times 3$ windows.

\subsection{Ablations}

	

The following ablation studies investigate the impact of three key components on overall tracking performance.



\paragraph{Positional Encoding} As reported in Table~\ref{table:pos_encoding}, the positional encoding strategy has notable impact on downstream task performance. 
\begin{table}[ht]
    \centering
    \caption{Effect of different positional encoding incorporated in the downstream task.}
    \begin{tabular}{c|c|c|c|c}
        Positional Encoding & Median & Mean & Std & Max \\ \hline
        Naive & 1.47 & 2.51 & 3.43 & 36.24 \\ 
        Learnable & 1.37 & 1.86 & 1.54 & 11.22 \\ 
        Frame-aware (Ours) & \textbf{1.02} & \textbf{1.44} & \textbf{1.35} & \textbf{10.23} \\ \hline
    \end{tabular}
    \label{table:pos_encoding}
\end{table}
The naive positional encoding simply applies $1D$ sine-cosine positional encoding over all the patches, and hence loses the temporal information about the patches, resulting in unsatisfactory results. If learnable positional encoding is used, the temporal positions are still needed to be learnt leading to sub-optimal performance. Interpolating from the central patch positions of the pretrained frames (frame-aware positional encoding) gives the best results.

\paragraph{Masking Ratio} We further compare the performance of different intermediate frame masking ratios in Table \ref{table:mask_ratios}.
\begin{table}[ht]
    \centering
    \caption[Tracking performance with FIMAE trained with different intermediate frame masking ratios.]{Tracking performance with FIMAE trained with different intermediate frame masking ratios. i.e. masking ratio of $\Omega_{frame}$.}
    \begin{tabular}{c|c|c|c|c}
        Frame Masking ratio & Median & Mean & Std & Max \\ \hline
        95\% & 1.09 & 1.47 & \textbf{1.24} & 10.34 \\ 
        98\% & \textbf{1.02} & \textbf{1.44} & 1.35 & \textbf{10.23} \\ 
        100\% & 1.08 & 1.78 & 2.09 & 15.12 \\
        \hline
    \end{tabular}
    \label{table:mask_ratios}
\end{table}
Best results are obtained with an intermediate frame masking ratio of $98\%$. While results with $95\%$ are largely equivalent, there is a notable reduction in performance when the entire frame is masked, which may be due to the lack of patches and its relative positions information during pretraining.

\paragraph{Effect of initialization}
Recall that the first template crop during both training and inference was obtained from the initial catheter tip location and is not updated. We explore its impact in Table \ref{table:initeffect}. To assess its importance, we conduct two experiments. First, we dynamically update the initial template frame during inference, like the rest. Second, we introduce random noise (2 to 16 pixels) to the initial tip location instead of updating the template. Our findings highlight the crucial role of initialization in tracking. Updating the initial template frame worsens performance due to greater accumulated prediction errors over time compared to the original setup. Additionally, even small noise levels of 2 pixels can noticeably affect performance, increasing the maximum error by 5 pixels.  

\begin{table}[ht]
    \centering
    \caption[Significance of initialization in catheter tip tracking.]{Significance of initialization in catheter tip tracking: How the performance is affected if first template frame is updated or some noise is introduced to the initial tip coordinates.}
    \begin{tabular}{c|c|c|c|c|c}
        Upate First template & init noise ($\mp$ px) & Median & Mean & Std & Max \\ \hline
        \ding{51} & 0 & 1.17 & 1.90 & 2.51 & 24.55 \\
        \ding{55} & 16 & 1.53 & 2.44 & 3.18 & 25.42 \\
        \ding{55} & 8 & 1.45 & 1.94 & 2.25 & 26.45 \\
        \ding{55} & 4 & 1.13 & 1.69 & 2.07 & 20.72 \\
        \ding{55} & 2 & 1.05 & 1.55 & 1.60 & 15.36 \\
        \ding{55} & 0 & \textbf{1.02} & \textbf{1.44} & \textbf{1.35} & \textbf{10.23} \\ 
        \hline
    \end{tabular}
    \label{table:initeffect}
\end{table}
\paragraph{Modality Bias}

\begin{table}[ht]
\centering
    \caption[Performance variation across modalities based on modality-specific training.]{Performance variation across modalities based on modality-specific training.}
\begin{tabular}{cccccccccc}
Trained on  & \multicolumn{3}{c}{Fluoro} & \multicolumn{3}{c}{Angio} & \multicolumn{3}{c}{Devices} \\ \hline
            & mean   & median   & max    & mean   & median   & max   & mean   & median   & max     \\ \hline
Fluoro data & 1.44   & 0.84     & 10.54  & 4.15   & 2.36     & 22.96  & 6.58   & 4.62     & 19.47   \\
Angio data  & 1.41   & \textbf{0.75}     & 11.42  & \textbf{1.49}   & \textbf{1.14}     & \textbf{5.56}  & 2.80   & 0.99     & 22.55   \\
All data    & \textbf{1.24}   & \textbf{0.75}     & \textbf{10.23}  & 1.61   & 1.38     & 7.33  
& \textbf{1.54}   & \textbf{0.98}     & \textbf{4.27}  \\ \hline 
\end{tabular}
\label{table:modalitybreak}
\end{table}

The distribution between Angio and Fluoro varies to some degree in terms of dosage and presence of contrasted vessel structures. We remind the reader that in our training dataset the distribution of Angio:Fluoro sequences was 2098:216 of the total of 2314 sequences. Our objective in this study is to develop a model that exhibits strong performance across both modalities. We present the results of training on individual modalities compared to training on combined data in Table \ref{table:modalitybreak}. Our findings indicate that training solely on one modality results in suboptimal performance on the other modality. Notably, while training on Angio data yields an improvement in Angio performance, training exclusively on Fluoro data fails to enhance performance in Fluoro. We hypothesize that a possible reason for this effect is the imbalance of 2098:216 (Angio to Fluoro sequences), with the following effects: 
\begin{enumerate}
    \item 2098 Angio sequences is a large enough dataset to ensure good Angio performance when training on this data alone;
    \item 216 Fluoro sequences is too little to power the training of a large transformer model, leading to inferior results when training/testing on Fluoro only;
    \item transitioning from Angio to using all data for training has a negative effect on the Angio test performance - we hypothesize that adding the few Fluoro sequences to training increases the complexity of the training problem, as the distribution of Angio training cases is enhanced with the distribution of Fluoro cases, based on only 216 examples; and
    \item transitioning from Fluoro to using all data for training has a positive effect on the Fluoro test performance - we hypothesize this is because the 216 Fluoro sequences are complement with many more non-contrasted frames from all Angio sequences to substantially increase the dataset, and thereby improve performance.
\end{enumerate}
Furthermore, the challenges posed by device obstruction exhibit nuanced differences between fluoro and angio, contributing to reduced performance when the model is trained on a single modality.

\section{Conclusion}
In this study, we present Frame Interpolation Masked Autoencoder (FIMAE), a Masked Imaging Modeling (MIM) approach, which is introduced for the purpose of acquiring generalized features from a large unlabeled dataset containing more than 16 million interventional X-ray frames, with the objective of device tracking. FIMAE overcomes the limitation of tube masking as proposed in VideoMAE, and applies frame interpolation-based masking for capturing fine inter-frame correspondences. The acquired features are subsequently applied to the task of device tracking within fluoroscopy and angiography image sequences. Our pre-trained FIMAE encoder surpasses all prevalent MIM-based pretraining methods for sequential imaging processing.

The spatio-temporal features acquired during the pretraining phase significantly influence the extraction and matching of features for the purpose of device tracking. We demonstrate that an efficient spatio-temporal encoder can replace the frequently utilized Siamese-like architecture, yielding a computationally lightweight model that maintains a high degree of precision and robustness in the tracking task. By adopting our methodology, we achieve a noteworthy 23.2\% reduction in maximum tracking error, even without the incorporation of supplementary modules such as flow refinement, when compared to the state-of-the-art multi-modular optimized approach. This performance enhancement is accompanied by a frame-level TSUC score of 97.95\% at $3 \times$ faster inference speed than the state-of-the-art method. The results also show that our approach achieves superior tracking performance, particularly in the challenging cases where occlusions and distractors are present.

\paragraph{Limitations and Future Work} Our investigation is primarily centered on leveraging pre-trained features for the tracking of devices within X-ray sequences. Consequently, we contend that the pre-trained model can be further extended to other tasks within interventional image analytics, such as stenosis detection, guidewire localization, and vessel segmentation. 
Furthermore, the absence of annotated frames within our sequential imaging dataset imposes a constraint on the utilization of historical trajectory information, a commonly exploited approach in recent single object tracking methodologies in the natural imaging domain. Thus, a more comprehensive investigation is needed to effectively make use of this information in our specific context.

\subsection*{Disclaimer}
The concepts and information presented in this paper are based on research results that are not commercially available.

\subsection*{Disclosures}
There are no conflicts of interest.

\subsection*{Code, Data, and Materials Availability}
Based on the data usage agreements, the data cannot be shared with the community. More information about the code can be shared upon request. 


\bibliography{report}   
\bibliographystyle{spiejour}   

\pagebreak
\appendix    

\section{Pretraining details}
The detailed architecture illustration and the implementation details of the pretraining are illustrated in Table \ref{tab:fimae_arch} and Table \ref{tab:fimae_impl} respectively. We use a 10-frame vanilla ViT-Base as our encoder architecture which incorporates joint space-time attention on visible patches. The decoder is of lower dimension and lower depth than encoder which incorporates similar joint space-time attention on all patches. The decoder is only responsible for reconstruction and is discarded for downstream tasks.

\begin{table}[!ht]
    \centering
    \begin{tabular}{c|c|c}
    \hline
        \textbf{Stage} & \textbf{Vision Transformer (Base)} & \textbf{Output Size} \\ \hline
        data & temporal gaps = [1,2,3,4] & 1×10×384×384 \\ \hline
        Patch Embed & 1×16×16, 768 & 768×10×576 \\
        ~ & stride 1×16×16 & ~ \\ \hline
        mask & $\rho$ = Tube 75\% + frame 98\% & 768×10×[576×(1-$\rho$)] \\ \hline
        encoder & [MHA(768), MLP(3072)] x 12 & 768×10×[576×(1-$\rho$)] \\ \hline
        projector & MLP(384) \& & ~ \\ 
        ~ & concat learnable tokens & 768×10×576 \\ \hline
        decoder & [MHA(384), MLP(1536)] x 4 & 384×10×576 \\ \hline
        projector & MLP(256)  & 256×10x576 \\ \hline
        reshape & from 256 to 1×1×16×16 & 1×10×384×384 \\ \hline
    \end{tabular}
    \caption{\textbf{Architecture details of FIMAE}. We use a 10-frame vanilla ViT-Base as our architecture. "MHA" here denotes the joint space-time self-attention. The output sizes are denoted by {C×T×S} for channel, temporal and spatial sizes.}
     \label{tab:fimae_arch}
\end{table}

\begin{table}[!ht]
    \centering
    \begin{tabular}{c|c}
    \hline
        \textbf{Config} & \textbf{Name/Params} \\ \hline
        Optimizer & AdamW \\ \hline
        Base learning rate & 1.5e-4 \\ \hline
        weight decay & 1e-4 \\ \hline
        optimizer momentum & $\beta_1$, $\beta_2$ = 0.9, 0.95 \\ \hline
        batch size & 8 \\ \hline
        learning rate schedule & cosine decay \\ \hline
        warmup epochs & 15 \\ \hline
        augmentation & MultiScaleCrop \\ \hline
    \end{tabular}
    \caption{\textbf{Pretraining Setting}}
    \label{tab:fimae_impl}
\end{table}

\pagebreak
\section{Downstream Model Details}
The architectural detail of the downstream tracking model is depicted in Table \ref{tab:track_arch}. The encoder is same as the pretraining encoder, whereas the decoder is a lightweight Transformer decoder, followed by two CNN heads that outputs the catheter tip heatmap and catheter body mask respectively. The implementation details are further explained in Table \ref{tab:track_impl}.

\begin{table}[!ht]
    \centering
    \begin{tabular}{c|c|c}
    \hline
        \textbf{Stage} & \textbf{ViT-Base + Multi-task Decoder} & \textbf{Output Size} \\ \hline
        data & 3 Templates + 1 Search & 1×3×64×64, 1×1×160×160 \\ \hline
        Patch Embed & 1×16×16, 768 & 768×148 \\
        ~ & stride 1×16×16 & ~ \\ 
        ~ & \& concatenate & ~ \\ \hline
        encoder & [MHA(768), MLP(3072)] x 12 & 768×148 \\ \hline
        projector & MLP(256) & 256×148 \\ \hline
        
        decoder & Query = (256, 2) \& & 256×148, 256×2 \\ 
        ~ & [CA(256), MLP(1024)] x 6 & ~ \\ \hline
        cross-correlate & Unconcatenate and matmul & 256×2×10×10  \\ \hline
        mask head & (Upconv, Conv, Batchnorm, GeLU) x 4  & 4×160×160 \\ \hline
        mask project & Linear(4,1) & 1×160×160 \\ \hline
        heatmap head & (Upconv, Conv, Batchnorm, GeLU) x 4  & 4×160×160 \\ \hline
        heatmap project & Linear(4,1) & 1×160×160 \\ \hline
    \end{tabular}
    \caption{\textbf{Architecture details of downstream tracking model}. "CA" refers to Cross-Attention.}
    \label{tab:track_arch}
\end{table}

\begin{table}[!ht]
    \centering
    \begin{tabular}{c|c}
    \hline
        \textbf{Config} & \textbf{Name/Params} \\ \hline
        Optimizer & AdamW \\ \hline
        Base learning rate & 6e-4 \\ \hline
        weight decay & 1e-4 \\ \hline
        optimizer momentum & $\beta_1$, $\beta_2$ = 0.9, 0.95 \\ \hline
        batch size & 42 \\ \hline
        learning rate schedule & cosine decay \\ \hline
        warmup epochs & 10 \\ \hline
        augmentation & Horizontal Flip \\
        ~ & Vertical Flip \\
        ~ & Random Rotation (-10$^{\circ}$, 10$^{\circ}$)  \\ \hline
    \end{tabular}
    \caption{\textbf{Finetuning Setting}}
    \label{tab:track_impl}
\end{table}

\pagebreak

\vspace{2ex}\noindent\textbf{Saahil Islam} is a second-year PhD student at Friedrich Alexander University in Erlangen, Germany. Having obtained an MSc from the same institution with a specialization in computer vision and segmentation for glacier calving front detection from remote sensing images, Saahil's current research focuses on medical imaging within the realm of image-guided therapy, conducted in collaboration with Siemens Healthineers. Specifically, Saahil is dedicated to leveraging artificial intelligence to enhance real-time systems in image-guided therapy, aiming to contribute to advancements in this critical field.

\vspace{2ex}\noindent\textbf{Venkatesh N. Murthy} a seasoned computer scientist, brings over a decade of experience in Computer Vision and Machine Learning. Having earned his Ph.D. from UMass Amherst, Venkatesh has garnered acclaim through numerous publications in prestigious conferences and journals, amassing over 700 citations and securing multiple patents. Currently, as a Staff Research Scientist at Siemens Healthineers in Princeton, NJ, Venkatesh focuses he focuses on advancing object classification, detection, and tracking technologies, driving innovation in healthcare technology.

\vspace{1ex}
\noindent Biographies and photographs of the other authors are not available.

\vspace{2ex}\noindent

\listoffigures
\listoftables

\end{spacing}
\end{document}